# An Efficient Technique for Image Captioning using Deep Neural Network

Author
Borneel Bikash Phukan
*School of Computer Engineering*
*KIIT University*
*Bhubaneswar, India*
borneelphukan@gmail.com

Co-Author
Dr. Amiya Ranjan Panda
*School of Computer Engineering*
*KIIT University*
*Bhubaneswar, India*
amiya.pandafcs@kiit.ac.in

*Abstract*—With the huge expansion of internet and trillions of gigabytes of data generated every single day, the needs for the development of various tools has become mandatory in order to maintain system adaptability to rapid changes. One of these tools is known as Image Captioning. Every entity in internet must be properly identified and managed and therefore in the case of image data, automatic captioning for identification is required. Similarly, content generation for missing labels, image classification and artificial languages all requires the process of Image Captioning. This paper discusses an efficient and unique way to perform automatic image captioning on individual image and discusses strategies to improve its performances and functionalities.

## I. INTRODUCTION

Computers are known for their capability to conduct high speed computations as compared to human beings. With the development of better processing powers, computers have also gained the power to compute complex mathematics to perform tasks just like the human mind. For examples, computers have the ability to use algorithms that can differentiate objects, perform speech recognition, face recognition, emotion recognition and much more. Proper combination and coordination between these algorithms have the ability to mimic the subconscious human mind with better performance than the normal human brain.

One such important algorithm or ability is to recognize an image entity and automatically generate a caption for it. This ability is called 'Automatic Image Captioning'. Image captioning is not new to the world of computer science. Multiple approaches has been made to conduct automatic image captioning, but the performance has been highly compromised due to the lack of computer resources and improper approach and optimization to the algorithm. Some models tend to use up more resources, which is a major drawback whereas some tends to use up more space. However, with proper combination of deep learning methodologies and optimization, these problems can be thwarted. Hence, the goal of this paper is to describe an efficient method to automatically generate caption for an image using deep neural network approach.

## 2. LITERATURE REVIEW

It is worth mentioning that intuition and development of image captioning technology is not new. Multiple corporates, research communities and open source communities have predicting image features from images and classifying the features. Since we are not classifying features but only been constantly striving to achieve higher performance and resilience in the technology. Since the advent of deep neural networks the race for a better performance has increased exponentially. Researchers developed multiple neural networks each with better performances and features than its predecessor.

Krizhevsky et al. [1] developed a neural network powered by GPU training procedures. The model was successful in drastically reducing the input dimension to produce (None, 1000) output without model overfitting. It uses 5 convolution layers backed by Maxpooling layers and proper implementation of dropout regularization which enhanced its performance. Karpathy and FeiFei [2][9] were the first to process image datasets and their corresponding sentence descriptions to enable computers generate image descriptions. It introduces a Multimodal Recurrent Neural Network (m-RNN) that uses the co-linear arrangement of features in order to carry out the task. Vinyals et al. [3] developed a generative model that consisted of an RNN which boosted machine translation and computer vision for image captioning by ensuring better probability of the generated sentence to accurately describe the target image. Xu et al. [4][11] developed an attention-based model that has the ability to automatically learn and describe an image. The model was trained using a standard backpropagation technique and it was able to identify object entities in image and simultaneously generate an accurate caption.

Using the bits of knowledge from the above mentioned works, we shall proceed with developing the proposed model. Every time we use a deep neural network, the model needs to be trained and tested using relevant dataset made for that particular purpose. The approach begins with the preprocessing of data. The preprocessing steps include feature extraction from image data, cleaning and tokenization of text data (if available) and make the entire data readable and compatible for the deep neural network model. Next task is to create a model that produces efficient state of the art performance on outcomes and minimize loss. A neural network might require huge number of optimization. This includes configuring the learning rate, the decay parameters, number of hidden layers etc. It may even include changing the whole architecture. For this paper, it is note mentioning that we will be using VGG16 neural network architecture for feature extraction from images and 'merge model' neural network architecture for training the entire image captioning system.

The VGG16 model for feature extraction must need re-configuration as the default model is used for
extracting the features for processing through the merge model, the removal of the last layer (classification layer)

from the VGG16 model is required. The architecture of the VGG16 model is given below:

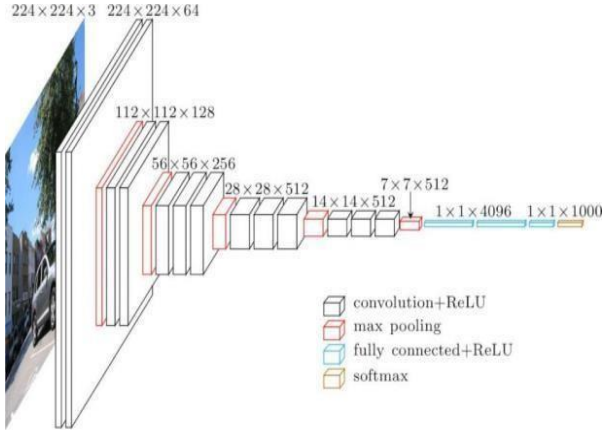

*Figure 1: VGG16 Architecture*

Following this, we will be designing a merge model neural network [5] and fitting the model with the preprocessed data and perform the model training and also subsequently keep track of the loss function output. Since training of merge model requires huge amount of RAM, and our goal is to reduce the consumption of resources, the training needs to be done through 'Progressive Loading' technique. Once the training is completed with minimum loss outcome, we proceed with model evaluation by calculating the BLEU performance of the model. Upon getting the desired performance, the model can be used to generate desirable captions for any individual image data the computer needs to identify.

## 3. DATA PREPARATION

We begin by preprocessing the data required for giving our model the ability to identify image. For this, we are using the Flickr8K dataset available in Kaggle [14]. The dataset is a collection of training, validation and test images all stored in JPEG format. The dataset is divided into two sub-parts. Flickr8_Dataset and Flickr8k_text. Flickr8_Dataset contains all the image data that is required for training, validation and testing and the Flickr8_text contains all the meta-data that determines the category of images and tokens for training the model in captioning. The structure of the dataset is given below:

*Table I: Flickr8k Dataset*

| Flickr8k Image Dataset (Flickr8k_Dataset) | |
|---|---|
| Total Images | 8092 |
| Training Data (images) | 6000 |
| Validation Data (images) | 1000 |
| Testing Data (images) | 1000 |
| **Flickr8k Text Dataset (Flickr8k_text)** | |
| Flickr8k token (text) | 40460 |
| Flickr8k lemmatized token (text) | 40460 |

First the content of each image should be interpreted by the computer. Hence the features of each and every images must be extracted to form feature vectors. To do this, we make use of the VGG16 Neural Network Architecture [6]. Like any other neural networks, VGG16 requires training to gain valuable insights from the image data. However since this consumes huge amount of time, it is suggested to import a pre-trained model to pre-compute the image features from the image data. It is note mentioning that the last layer of the neural network must be removed as it is used for classification of predicted features.

Following this, we have to preprocess the text data available as Flickr8k.token.txt. This file contains all the text data which the model shall use to understand the language for image captioning. Using natural language processing techniques, the text data are tokenized and separated into image id and description. The description are then cleaned. The cleaning procedure mostly involves conversion to lower case, removing numbers from text, apostrophe 's', 'a' and punctuation. These cleaned descriptions along with their image id are then stored as a text file for further use. This storage file consisting of the processed text data is known as text embeddings file.

## 4. NEURAL NETWORK ARCHITECTURE

In the data preparation section, we have preprocessed two entities of data. Image data and text (caption) data to produce encoded image feature vectors and encoded text embeddings respectively. Both the preprocessed data should be able to fit in the neural network. However, image feature vectors and text embeddings together cannot be fitted to the same input layer of a single neural network. Therefore, it is necessary that there are separate input layers for the two input types of data. Henceforth the concept of 'merge model' comes into play.

### 1. Merge Model

In a merge model, two neural networks combine the two types of encoded data input which is then used by a simpler decoder network to generate the next word in the sequence of words in the caption. The preprocessed text needs to be passed through an embedding layer followed by a recurrent neural network known as LSTM (Long Short Term Memory)[8]. The image feature vectors on the other hand passes through a densely connected neural network layer followed by feed forward network ending with a softmax layer. It is in this second dense layer that the encoded outputs from the LSTM in first neural network merges with the encoded image embeddings output from the second neural network and decodes to predict the next word in the sequence of words of caption.

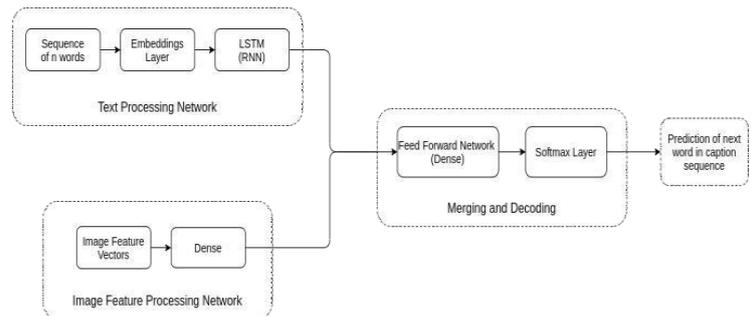

*Figure 2: Merge Model*

### 2. Long Short Term Memory (LSTM)

Since in the first network we are processing textual data, we are performing Natural Language Processing (NLP) and for NLP, the most favourable neural network is considered to be a Recurrent Neural Network (RNN) [12]. And from the group of RNN, due to the advantages of overcoming vanishing and exploding gradient problem, it is favourable to use the Long Short Term Memory (LSTM) network [7]. LSTM has the ability to back-propagate the error through unlimited number of time steps. It has three gates: input, forget and output gates. These gates calculate the hidden state by taking their combination according to the equations given below[7]:

$$\mathbf{x} = \begin{bmatrix} \boldsymbol{h}_{t-1} \\ \boldsymbol{x}_t \end{bmatrix}$$
$$f_t = \sigma(W_f \cdot \mathbf{x} + b_f)$$
$$i_t = \sigma(W_i \cdot \mathbf{x} + b_i)$$
$$o_t = \sigma(W_o \cdot \mathbf{x} + b_o)$$
$$c_t = f_t \odot c_{t-1} + i_t \odot tanh(W_c \cdot X + b_c)$$
$$h_t = o_t \odot tanh(c_t)$$

3. Model Designing

Putting all the elements together in the merge model, we shall now descriptively design the neural network architecture.

The image feature processing neural network inputs 1D image embeddings of dimension 4096 followed by a regularization layer with 50% dropout to prevent overfitting of the model during training due to rapid learning procedure. Following this, it is passed through a dense layer which reduces the 1D vector dimension from 4096 to 256 and then into the addition layer.

The text processing neural network handles the 1D text input vector of dimension 34 and passes it through an embedding layer which converts the words into word embeddings and again perform regularization with 50% dropout to counter overfitting while training. Following this, it is passed through an LSTM network of same dimension to help generate semantically rich and descriptive sentence for a given image.

Upon receiving the outcomes from the two neural networks, we create the merging and decoding section of the neural network. This layer includes the addition layer which merges the two outcomes, followed by a dense 1D layer of dimension 256 having 'relu' activation and another dense layer (or output layer) of same dimension with 'softmax' activation.

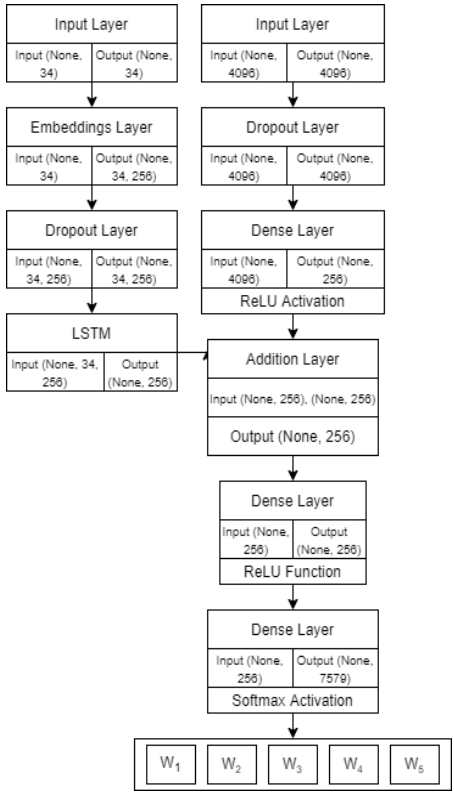

*Figure 3: Merge Model Network*

## 5. MODEL TRAINING

After designing of the model and preparing and preprocessing of the data, the next step is to fit the data with the model and iteratively run it on the training data. The training procedure requires a lot of computer resources as it involves iterating across the entire training dataset multiple times and calculating the loss function. Therefore rather than loading the entire dataset all at once, we have devised a new method that gradually loads the data by means of a generator function. This is known as Progressive Loading.

We know that the size of a dataset is always proportional to the memory requirement. But as we are designing an efficient method, in order to overcome this, we create a generator function called data generator. This generator function converts an input array consisting of image embeddings and encoded word sequences to sequentially yield one-hot encoded words. These one-hot encoded words along with the image embeddings are progressively fed into the two sets of neural networks and sequentially trained to generate the desirable weights. Hence the name Progressive Loading.

Following loading of data, we proceed with the model training. Here we are using 20 epochs, each epochs consisting of 6000 training images. It is note mentioning that the number of epochs has been taken on an intuitive basis and fortunately these number of epochs has shown to drastically reduce the model loss, also preventing underfitting and overfitting of the model. Also, since model training traditionally takes a certain amount of time, it is worth advisable to use model checkpoints which periodically saves the model after completion of each epoch.

## 6. MODEL EVALUATION

Once the model fitting and training has been completed, a model evaluation procedure ensures the integrity of its prediction. For model evaluation the model is run using the validation dataset and the performance is monitored by the ability of the model to generate a caption word-by-word after proper mapping with the trained image feature vectors. Once the caption generation is done, we need to compare and evaluate the difference between the actual output and the predicted output over the validation dataset. For the evaluation of predicted caption output, there is a need to use an evaluation metric relevant for textual data evaluation. In the context of this paper we shall be using BLEU evaluation metric for the same [8].

Bilingual Evaluation Understudy Score (BLEU) is one of the most commonly used evaluation metric system for evaluating generated textual data with a given textual data [10]. Its range lies from 0 to 100, representing perfect mismatch and perfect match respectively. The metric has the advantage of comparing n-grams of the predicted text output with the n- grams of the reference data and can return the number of matches found. Hence higher the number of matches, better the model performance. The mathematics involved in the derivation of the BLEU formula is out of the scope of this paper, however following is an abstract glimpse of the derivation as described in Papineni K. et al [8].

We compute the length of the predicted (candidate) data and the validation (reference) data (denoted by c and r respectively) to determine the brevity penalty (BP).

$$BP = \begin{cases} 1 & \text{if } c > r \\ e^{(1-r/c)} & \text{if } c \leq r \end{cases}.$$

Then using the computed geometric average of the modified n-gram precisions, $p_n$, uniform weights $w_n$,

and Brevity Penalty (BP), we compute the BLEU using the formula,

$$\text{BLEU} = \text{BP} \cdot \exp\left(\sum_{n=1}^{N} w_n \log p_n\right)$$

On smoothening, we derive

$$\log \text{BLEU} = \min(1 - \frac{r}{c}, 0) + \sum_{n=1}^{N} w_n \log p_n.$$

For the context of this paper, we are using 4-gram scores each with different weights.

## 7. RESULTS

The primary focus while training the data is to ensure that the value of the loss function does not increase rapidly or remain stagnant with each epoch but display a gradual decrease. In order to monitor the performance, we specially monitor the training loss.

### A) Training Loss

We have trained the model using 6000 image data with 20 epochs. For each epoch, we noted the corresponding loss value and it was found to reduce with each epoch. The plot for model loss with each epoch is given below:

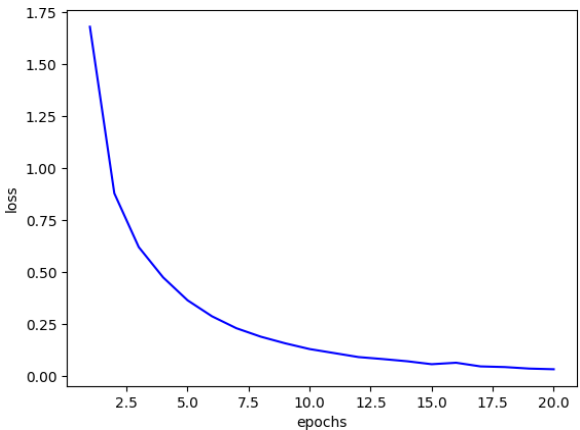

*Figure 5: Epoch vs Loss Curve*

### B) BLEU Performance

Following the model training we determine the accuracy of the model to generate the respective caption. As discussed in the section, we are using BLEU metric to determine the accuracy of word generation. We are testing the text generation procedure using 4-gram scores. Each gram corresponds to different weights. Following are the observations made on performing BLEU evaluation.

*Table 2: BLEU Score Analysis*

| N-GRAM | WEIGHTS | SCORE |
|--------|---------|-------|
| BLEU-1 | 1.0, 1.0, 1.0, 1.0 | 79.0190 |
| BLEU-2 | 0.5, 0.5, 0, 0 | 52.8925 |
| BLEU-3 | 0.3, 0.3, 0.3, 0 | 37.4354 |
| BLEU-4 | 0.25, 0.25, 0.25, 0.25 | 18.3087 |

### C) Testing on a random image

Once the model has been trained and evaluated, we ran the model on a random .PNG image. The model has been found to efficiently identify the image and caption it accordingly. These captions are stored as text data. However it should be noted that the model is not trained to identify people, landmark and signboards. However with customization of the model, the model can be made to identify various entities.

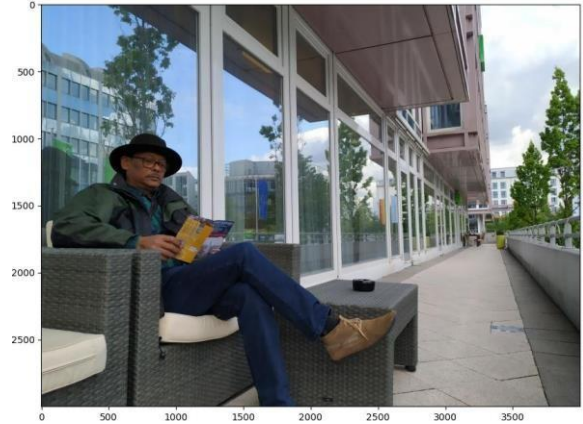

*Figure 7: Test Image 1*

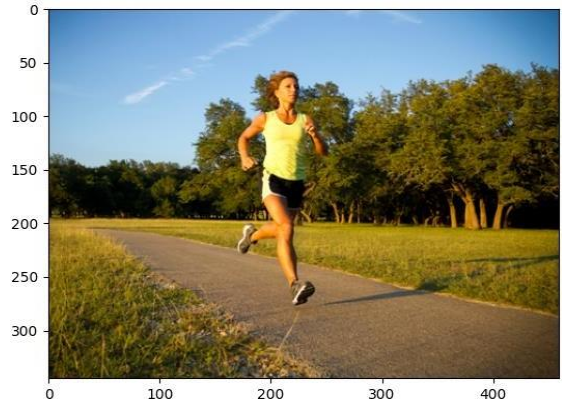

*Figure 8: Test Image 2*

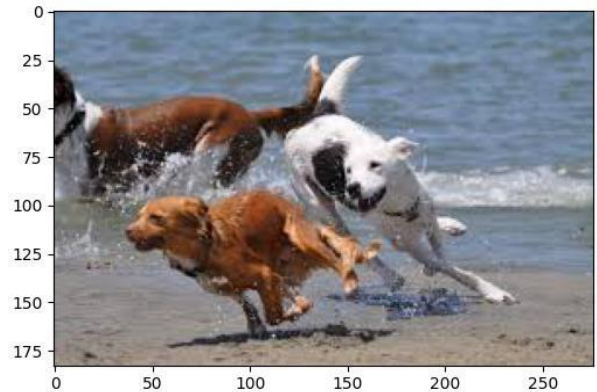

*Figure 9: Test Image 3*

```
borneel@borneel-ubuntu:~/Desktop/Research Paper$ python3 generate.py
[INFO] [Image 1]: Generated Caption: A man is sitting on chair
[INFO] [Image 2]: Generated Caption: Dog playing in water
[INFO] [Image 3]: Generated Caption: A man is running
```

*Figure 10: Model Outputs*

## 8. CONCLUSION

In this paper, we have introduced a new and efficient method for carrying out automatic caption generation. We have used Flickr8k dataset for extracting image features using VGG16 neural network, preprocessed textual caption data, designed a merge model neural network that merged the extracted image features with text data together to sequentially generate probable words for captioning the images. We have trained the merge model on 6000 training data. Furthermore, in order to reduce computational load while loading the data into the model, we have introduced a new loading method known as Progressive Loading, which makes use of a generator function to step by step yield data into the model. To monitor the performance of the caption generation and generated loss of the model while training, we have used a BLEU evaluation method and categorical cross entropy loss function respectively. After following every procedures, the model was tested using a random picture and was found to generate the caption with a perfect BLEU score.

## 9. REFERENCES

[1]. Krizhevsky A., Sutskever I., and Geoffrey E. Hinton, ImageNet Classification with Deep Convolutional Neural Networks, Communication of the ACM, May 2017 Vol. 60 No. 6

[2]. Karpathy A., Fei-Fei L., Deep Visual Semantic Alignments for Generating Image Descriptions, IEEE Transactions on Pattern Analysis and Machine Learning, 1 April 2017, Page(s): 664 -676 Electronic ISSN: 1939-3539 Vol. 39 Issue 4

[3]. Vinyals O., Toshev A., Bengio S., Erhan D., Show and Tell: A Neural Image Caption Generator, IEEE Conference on Computer Vision and Pattern Recognition (CVPR), 15 October 2015, Print ISSN:1063-6919

[4]. Kelvin Xu, Jimmy Lei Ba, Ryan Kiros, Kyunghyun Cho, Aaron Courville, Ruslan Salakhutdinov, Richard S. Zemel, Yoshua Bengio, Show, Attend and Tell: Neural Image Caption Generation with Visual Attention, [Online] Available: https://arxiv.org/pdf/1502.03044.pdf

[5]. Chou Y. M., Chan Y. M., Lee J. H., Chiu C. Y. and Chen C. S. "Unifying and Merging Well-trained Deep Neural Networks for Inference Stage", Twenty-Seventh International Joint Conference on Artificial Intelligence, July 2018, Page(s): 2049-2056.

[6]. Zhang X., Zou J., He K. and Sun J. "Accelerating Very Deep Convolutional Networks for Classification and Detection", IEEE Transactions on Pattern Analysis and Machine Intelligence, 1 October 2016, Page(s) 1943-1955, Electronic ISSN-0162-8828

[7]. Greff K., Srivastava R. K., Koutník J., Steunebrink B. R., and Schmidhuber J. "LSTM: A Search Space Odyssey", IEEE Transactions on Neural Networks and Learning Systems, 8 July 2016, Page(s): 2222-2232, Electronic ISSN: 2162-2388

[8]. Papineni K., Roukos S., Ward T., Zhu W. J., "BLEU: a Method for Automatic Evaluation of Machine Translation", Proceedings of the 40th Annual Meeting of the Association for Computational Linguistics, July 2002, Page(s): 311-318.

[9]. Mao J., Xu W., Yang Y., Wang J., Huang Z., and Yuille A., Deep Captioning with Multimodal Recurrent Neural Networks (m-RNN). In International Conference on Learning Representations (ICLR).

[10]. Hossain MD. Z., Sohel F., Shiratuddin M. F., Laga H., "A Comprehensive Survey of Deep Learning for Image Captioning", "ACM Computing Surveys", February, 2019, vol. 51, Issue 6, 118.

[11]. Yang Z., Zhang Y. J., Rehman S., Huang Y., "Image Captioning with Object Detection and Localization", International Confernece on Image and Graphics (ICIG), 2017,

[12]. Shubo Ma and Yahong Han. 2016. Describing images by feeding LSTM with structural words. In Multimedia and Expo (ICME), 2016 IEEE International Conference on. IEEE, 1–6.

[13]. Karen Simonyan and Andrew Zisserman. 2015. Very deep convolutional networks for large-scale image recognition. In International Conference on Learning Representations (ICLR).

[14]. Flickr8k Image Captioning Dataset from Kaggle – [Available] – kaggle.com/shadabhussain/automated-image-captioning-flickr8